\documentclass[letterpaper, 10 pt, conference]{ieeeconf}  %

\IEEEoverridecommandlockouts                              %

\usepackage{amsmath}
\usepackage{amssymb}
\usepackage{graphicx}
\usepackage{url}

\title{\LARGE \bf
TacZero: Training-Free Peg Insertion Using a General-Purpose Vision-Language Model with Tactile Feedback
}

\author{Kazutoshi Tanaka$^{1}$%
\thanks{$^{1}$OMRON SINIC X Corporation, Bunkyo-ku, Tokyo 113-0033, Japan
        {\tt\small kazutoshi.tanaka@sinicx.com}}%
\thanks{This work has been submitted to the IEEE for possible publication.
Copyright may be transferred without notice, after which this version may
no longer be accessible.}%
}

\begin{document}
\bstctlcite{taczero-bib-style}

\maketitle
\thispagestyle{empty}
\pagestyle{empty}

\begin{abstract}

Robots that autonomously determine their actions from language instructions and sensory observations could perform new contact-rich manipulation tasks without task-specific training or hand-designed rules.
To perform these tasks, robots must infer how objects contact one another and move as a result, then select actions.
For contact inference and action selection, prior approaches involve designing estimation models and tactile feedback control laws, or learning models for object-motion estimation, action-outcome prediction, and action selection from tactile data.
Instead, we propose TacZero, which uses a pretrained general-purpose vision-language model (VLM) to interpret visual and tactile observations and select robot actions without additional tactile or manipulation training or task-specific rules for contact interpretation or action selection.
TacZero provides the VLM with camera images, robot state, and three-axis tactile responses represented as numerical values or vectors overlaid on the images.
From these observations and interaction history, the VLM generates commands specifying target end-effector positions and gripper opening or closing, which a low-level controller executes.
In real-world cylindrical-peg insertion experiments, TacZero succeeded in 15 of 20 trials with numerical tactile input, compared with 10 of 20 without tactile input.
This study provides a concrete starting point for further research on contact-rich manipulation using general-purpose VLMs and highlights challenges in pursuing this direction.

\end{abstract}

\section{INTRODUCTION}

Robots that autonomously determine their actions from language instructions and sensory observations could perform new contact-rich manipulation tasks~\cite{suomalainen2022survey,tang2023industreal}, such as peg insertion~\cite{kim2021active}, without task-specific training or hand-designed rules.
Contact-state estimation and reasoning about how contact constrains object motion can guide action selection in these tasks~\cite{kim2023simultaneous,cheng2022contact}.
Tactile sensing supports contact-state estimation by providing information about contact that is difficult to obtain from vision alone~\cite{li2020tactileinformation}.

\begin{figure}[t]
  \centering
  \includegraphics[width=\linewidth]{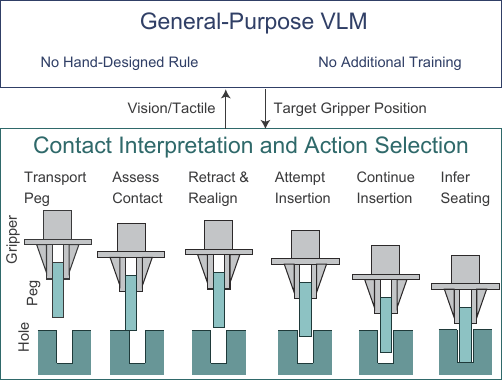}
  \caption{TacZero uses a pretrained general-purpose VLM to interpret visual and tactile observations and select robot actions during peg insertion, without additional tactile or manipulation training or hand-designed task-specific rules for contact interpretation or action selection.}
  \label{fig:teaser}
\end{figure}

Prior work has developed model-based and learning-based approaches to contact-state estimation and action selection.
Model-based approaches combine tactile measurements with geometric and contact constraints to estimate contact states~\cite{kim2021active} and determine robot motions through designed tactile feedback control laws~\cite{li2013tactileservoing}.
Learning-based approaches train models to estimate object motion~\cite{kim2021active} or predict action outcomes from tactile data~\cite{tian2019manipulation}, or learn policies to select actions~\cite{dong2021tactilerl}.
Other studies learn tactile representations that can be reused across perception tasks~\cite{higuera2025sparsh} or incorporate tactile feedback into pretrained robot policies through fine-tuning on demonstrations from multiple manipulation tasks~\cite{jones2025beyondsight}.
These approaches enable tactile-based manipulation through the design of models and control laws for contact-state estimation or action selection, or through learning from tactile data.

Vision-language models (VLMs) have enabled robotic manipulation without task-specific training~\cite{nasiriany2024pivot,fang2024moka,huang2024copa}.
These models use visual observations and language instructions to generate task plans~\cite{wang2024wonderfulteam}, robot programs~\cite{zhi2025come}, and spatial targets or constraints for motion execution~\cite{fang2024moka,huang2024copa}.
These advances suggest that a pretrained general-purpose VLM could also interpret visual and tactile observations and select robot actions without additional tactile or manipulation training or hand-designed task-specific rules for contact interpretation or action selection.

To explore this possibility, we propose TacZero (Fig.~\ref{fig:teaser}), which uses a pretrained general-purpose VLM to interpret contact from visual and tactile observations and select robot actions, without additional tactile or manipulation training or hand-designed task-specific rules for contact interpretation or action selection.
At each step, TacZero provides the VLM with the task instruction, visual and tactile observations, robot state, and the within-trial interaction history.
The VLM outputs commands specifying target end-effector positions and gripper opening or closing, which a low-level controller executes.
In real-world cylindrical-peg insertion experiments, TacZero succeeded in 15 of 20 trials with numerical tactile input, compared with 10 of 20 without tactile input.

Our contributions are:
\begin{itemize}
  \item We present a research direction in which a pretrained general-purpose VLM interprets visual and tactile observations and selects robot actions without additional tactile or manipulation training or hand-designed task-specific rules for contact interpretation or action selection. We provide a concrete starting point for future research and highlight challenges in pursuing this direction.
  \item We introduce TacZero, which provides a general-purpose VLM with a task instruction, visual and tactile observations, robot state, and interaction history, and executes the VLM's output commands specifying target end-effector positions and gripper opening or closing.
\end{itemize}

\section{RELATED WORK}

\subsection{Robotic Manipulation with Pretrained Language Models and Vision-Language Models}

Researchers have proposed robotic manipulation methods in which pretrained language models (LMs) and VLMs generate policy code or action sequences before execution.
Liang et al. proposed Code as Policies (CaP), which employs an LM to generate programs that process perception outputs, compute positions and trajectories, and invoke control application programming interfaces (APIs)~\cite{liang2023code}.
Fang et al. proposed MOKA, in which a VLM selects grasping and interaction keypoints, pre-contact and post-contact waypoint regions, and motion attributes, which are then converted into robot motions using depth information and geometric processing~\cite{fang2024moka}.
Huang et al. proposed CoPa, in which a solver computes a target pose from VLM-generated spatial constraints between object parts, and the VLM also specifies subsequent numerical displacements and gripper opening~\cite{huang2024copa}.
Whereas these methods execute the generated programs or action sequences without calling the LM or VLM again for action selection during execution, TacZero calls the VLM after each action to determine the next action from fresh observations.

Other researchers have proposed methods that call LMs or VLMs again based on execution feedback to update plans or actions.
Zhi et al. proposed COME-robot, which generates code that calls exploration, navigation, grasping, and placement functions and revises its plans using images and execution feedback~\cite{zhi2025come}.
Kwon et al. proposed a method that generates code to compute end-effector pose sequences and gripper commands from object geometry, then supplies observed object motion after trajectory execution to the LM for failure detection and replanning~\cite{kwon2024trajectory}.
Wang et al. proposed Wonderful Team, in which VLM agents identify and refine target locations in images, specify action sequences, compute waypoints using generated functions for trajectory-based tasks, and support replanning based on execution results~\cite{wang2024wonderfulteam}.
Nasiriany et al. proposed PIVOT, which presents candidate relative end-effector translations on the current image at each action selection and iteratively updates the candidate distribution based on VLM selections~\cite{nasiriany2024pivot}.
Whereas these methods specify motions through operation-level functions, trajectory-generating code, image-space target coordinates, or motion-candidate selection, TacZero uses the VLM to specify the next end-effector target position numerically.
Singh et al. proposed MALMM, which distributes planning, code generation, and coordination across multiple LM agents and replans using environmental observations obtained after intermediate execution steps~\cite{singh2025malmm}.
In MALMM, the LM specifies numerical end-effector target positions within generated code; similarly, the VLM in TacZero determines numerical end-effector target positions.

None of the methods discussed in this subsection provide tactile observations to the LM or VLM.
In contrast, TacZero provides a general-purpose VLM with tactile observations in addition to vision, robot state, and interaction history.

\subsection{Tactile and Force Feedback for Robot Action Selection and Control}

Researchers have proposed methods that integrate tactile feedback into robot action selection and refinement through additional training.
Hao et al. proposed TLA, which uses sequential tactile images and learns corrective insertion actions from tactile-action-instruction data~\cite{hao2026tla}.
Zhang et al. proposed VTLA, a vision-language-action (VLA) model that uses supervised fine-tuning and preference learning to select corrective actions from visual and temporal tactile observations~\cite{zhang2026vtla}.
Bi et al. proposed VLA-Touch, which uses object-property descriptions from a pretrained tactile-language model for VLM-based planning and a trained tactile-conditioned controller to refine action sequences generated by a VLA.
Although they also evaluated planning with tactile images supplied directly to a VLM, their main architecture assigns tactile-to-language interpretation and action refinement to dedicated models~\cite{bi2026vlatouch}.
Van der Merwe et al. proposed TAP-VLA, which overlays tactile shear vectors onto camera images and fine-tunes a VLA on these images~\cite{vandermerwe2026tapvla}.
TacZero adopts a tactile overlay representation inspired by TAP-VLA. Unlike TAP-VLA, which fine-tunes a VLA on the overlaid images, TacZero supplies these images to a general-purpose VLM without additional training.
Whereas these approaches use models trained for tactile-based action selection or refinement, TacZero uses a single general-purpose VLM to interpret visual and tactile observations and select robot actions without additional tactile or manipulation training.

Other researchers have proposed methods that combine a VLM with force-based control and search procedures without additional training.
Yajima et al. proposed a method that employs a VLM to identify the mating hole and determine the relative orientation, and executes insertion using predefined force-based control and search procedures~\cite{yajima2025zeroshot}.
Whereas the VLM in their method identifies the mating hole and determines the relative orientation, leaving insertion motions to predefined control and search procedures, the VLM in TacZero determines the next end-effector target position from observations at each step.

\section{TacZero}

\subsection{Task Setting}

We consider robotic peg insertion specified by a natural-language instruction. The instruction describes the peg and the desired insertion outcome without specifying target coordinates or a sequence of actions. The robot must insert the peg into the corresponding hole and maintain its grasp until insertion is complete. We do not perform additional tactile or manipulation training or update the VLM's parameters.

The robot has access to camera images, robot state, and tactile responses along the X, Y, and Z axes of the sensor coordinate frame, measured by a sensor mounted on the inner surface of one gripper finger. The VLM is not provided with explicit dimensions of the peg or the hole. Given the instruction and the interaction history available up to the current step, the VLM specifies the next target position of the end effector and a gripper opening or closing command. The end-effector orientation remains fixed throughout VLM-controlled execution.

A trial comprises a sequence of observation and action steps. It ends when the VLM calls \texttt{report\_completion}, the experimenter stops execution, or a predefined execution limit or safety stop is triggered. An insertion is successful when the peg's lower end is near the bottom of the specified hole. The experimenter judges success through visual inspection, independently of the VLM's completion report. If the experimenter observes that the peg has slipped out of the grasp or judges that slippage within the grasp prevents further insertion to the required depth, they stop the trial and record it as a failure. No interaction history or human-assigned outcome labels from previous trials are provided to the VLM.

\begin{figure}[t]
  \centering
  \includegraphics[width=\linewidth]{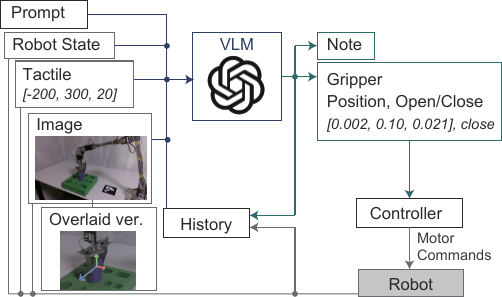}
  \caption{Overview of TacZero. Given the task instruction, visual and tactile observations, robot state, and interaction history, the VLM outputs a target end-effector position, a gripper opening or closing command, and a note containing a short observation and reason. The controller executes the command, and new observations are fed back to the VLM.}
  \label{fig:method}
\end{figure}

\subsection{Overview}

TacZero uses a pretrained general-purpose VLM to interpret visual and tactile observations and select robot actions, without additional tactile or manipulation training or hand-designed task-specific rules for contact interpretation or action selection. Fig.~\ref{fig:method} illustrates the overall framework. At each step, the VLM receives the task instruction, the current camera image, tactile responses, robot state, and the interaction history from the same trial. Tactile responses are presented as numerical values or vectors overlaid on the camera image. The VLM outputs the next end-effector target position and gripper opening or closing command, together with a note containing a short observation and reason. A low-level controller executes the command, and the resulting observations are supplied to the VLM at the next step. The interaction history includes observations, commands, notes, and execution results from the same trial, enabling the VLM to relate previous actions to subsequent changes in visual and tactile observations.

\subsection{VLM Inputs and Action Selection}

At each step, the VLM receives the task instruction, camera image, tactile responses, and robot state. The image provides information about the spatial relationships between objects and the end effector, while tactile responses provide complementary information about contact that may be difficult to discern from images alone. Robot state includes the end-effector position and orientation, joint and gripper encoder counts, and motor currents. We also provide camera calibration parameters and coordinate transformations to relate image observations to the robot reference frame. We instruct the VLM to specify target end-effector positions as absolute XYZ coordinates in this frame, expressed in meters, and provide the workspace bounds and execution limits. Observations also include acquisition timestamps.

We instruct the VLM to include a short observation and reason in the note accompanying the next end-effector target position and gripper opening or closing command. TacZero retains the command and its accompanying note in the within-trial interaction history together with subsequent observations and execution results, and provides this history to the VLM at the next step. This allows the VLM to relate previous actions and their stated reasons to subsequent changes in visual and tactile observations when selecting the next action. This interleaving of verbal reasoning, actions, and observations is similar to that used in ReAct~\cite{yao2023react}.

\begin{figure}[t]
  \centering
  \includegraphics[width=\linewidth]{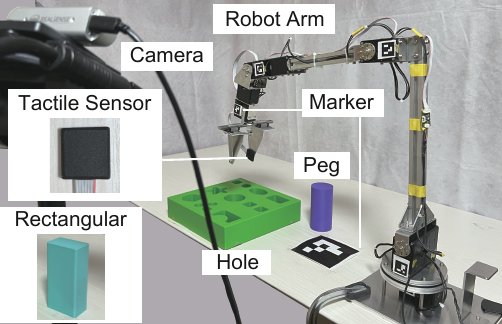}
  \caption{Peg-insertion setup with a robot arm, an overhead camera, and a tactile sensor mounted on one gripper finger. Insets show the tactile sensor and rectangular-prism peg.}
  \label{fig:setup}
\end{figure}

\subsection{Implementation Details}

Fig.~\ref{fig:setup} shows the experimental apparatus. We use a six-degree-of-freedom arm with the same joint-axis configuration, motors, and link lengths as the ALOHA follower arm~\cite{zhao2023aloha}. The arm is equipped with a two-finger gripper with flat inner contact surfaces, one of which serves as the mounting surface for a uSkin uSPa44 tactile sensor (XELA Robotics). The sensor consists of 16 taxels arranged in a $4\times4$ array, each measuring responses along the X, Y, and Z axes of the sensor coordinate frame. An overhead RGB-D camera (RealSense D435, Intel) observes the workspace. RGB images are supplied at a resolution of $1920\times1080$ pixels.

A stationary ArUco marker~\cite{garrido2014aruco} defines the reference coordinate frame. We estimate the pose of a hand-mounted ArUco marker from the camera image and transform it into the tool center point (TCP) pose in the reference frame using calibrated coordinate transformations. We use the TCP as the reference point for specifying target end-effector positions.

Before motor initialization and grasp preparation in each trial, we compute a baseline for each taxel as the component-wise median of 30 unloaded tactile frames. We keep this baseline fixed throughout the trial. Let $\mathbf s_{j,t}\in\mathbb R^3$ denote the XYZ response of taxel $j$ at time $t$, and let $\mathbf b_j\in\mathbb R^3$ denote its baseline. We compute the aggregate tactile response $\mathbf f_t^S \in \mathbb R^3$ in the sensor coordinate frame $S$ as
\begin{equation}
\mathbf f_t^S=\sum_{j=1}^{16}\left(\mathbf s_{j,t}-\mathbf b_j\right),
\end{equation}
where the sum includes all 16 taxels. We express these responses in sensor counts rather than calibrated force units. For each image, we select the latest tactile sample received at or before the image capture time.

We present the aggregate tactile response to the VLM in two forms: numerical values or vectors overlaid on the RGB image. In the numerical representation, we provide the three components of $\mathbf f_t^S$ alongside the unmodified RGB image. In the overlay representation, we render the signed responses along the sensor X, Y, and Z axes as red, green, and blue arrows, respectively. Let $E$, $0$, and $C$ denote the end-effector, stationary reference, and camera frames, respectively, and let $R_{AB}$ denote the rotation matrix that maps vector components from frame $B$ to frame $A$. For each sensor axis $i\in\{X,Y,Z\}$, let $f_{i,t}^S$ denote the corresponding component of $\mathbf f_t^S$. We transform each axis component into the camera frame as
\begin{equation}
\mathbf v_{i,t}^{C}=R_{C0}R_{0E,t}R_{ES}\left(f_{i,t}^{S}\mathbf e_i\right),
\end{equation}
where $\mathbf e_i$ is the unit vector along axis $i$ in the sensor frame. We determine $R_{ES}$ from the sensor mounting, obtain $R_{0E,t}$ from the measured end-effector orientation, and obtain $R_{C0}$ from the pose of the stationary reference marker in the camera frame.

We then project these camera-frame vectors onto the RGB image, using the projected TCP position as the origin of each arrow. Let $\mathbf p_{\mathrm{TCP},t}^{C}$ denote the measured TCP position in the camera frame, and let $\pi$ denote the camera projection function incorporating intrinsic parameters and distortion coefficients. We use a local linear approximation of the camera projection at the TCP position to map each three-dimensional tactile-response vector to a two-dimensional image arrow. We compute the two-dimensional displacement from the origin to the endpoint of each arrow, before applying length limits, as
\begin{equation}
\Delta\mathbf u_{i,t}=\alpha J_{\pi}\left(\mathbf p_{\mathrm{TCP},t}^{C}\right)\mathbf v_{i,t}^{C},
\end{equation}
where $\Delta\mathbf u_{i,t}$ is expressed in pixels and $J_{\pi}$ is the Jacobian of $\pi$ evaluated at the TCP position, computed using central differences. We use a fixed display scale of $\alpha=1.875\times10^{-5}$ m/count to convert tactile responses into lengths for visualization. We limit arrow lengths to 240 pixels and clip them at the image boundary. We inform the VLM whether any arrows have been clipped. The arrow origin is a display anchor, not an estimated contact location. Projection onto the image plane and arrow clipping can reduce the directional and magnitude information conveyed by the overlay, whereas the numerical representation preserves all three aggregate components without projection or clipping.

We use GPT-6 Astra (\texttt{gpt-6-astra}) as the primary VLM, accessed through the API with the reasoning effort set to \texttt{medium}. At each step, Astra receives the observations and interaction history and returns a structured tool call. The tool \texttt{move\_to(x, y, z, gripper, note)} accepts an absolute TCP target position in the reference frame, a gripper opening or closing command, and a note containing a short observation and reason. The VLM can specify each target coordinate as a numerical value in meters or as null. A null coordinate leaves the corresponding target coordinate unchanged. Before the first command, we initialize the target position from the measured TCP position. When all coordinates are null and a gripper command is specified, only the gripper is actuated. When a command specifies both translation and gripper opening or closing, the controller executes the translation first, followed by the gripper operation. Calling \texttt{report\_completion(note)} ends the trial and records the VLM's final note, which may report completion or state that the task remains incomplete.

During each movement, the controller repeatedly uses the marker-based TCP pose to update the position and orientation errors relative to the fixed target. It computes translational and angular velocity commands from these errors and converts them into joint velocity commands through inverse differential kinematics. The translational velocity command is proportional to the position error and capped at 20 mm/s, decreasing as the TCP approaches the target. The target end-effector orientation is fixed so that the end effector points downward relative to the reference frame. Each movement runs for the initial TCP-to-target distance divided by 20 mm/s, plus 3 seconds, with timing starting when the first velocity-command transmission completes. Reaching this time limit ends the movement regardless of whether the target has been attained. The controller returns execution results to the VLM, including execution duration, whether the time limit was reached, whether the target was attained, and final position and orientation errors when available. The VLM receives these results together with fresh observations for selecting the next action. We limit execution under VLM control to 100 action commands, 600 s of accumulated actuator-invocation time, or 1800 s of wall time, whichever is reached first.

\section{EXPERIMENTS}

\subsection{Setup}
\label{sec:experimental-setup}

We evaluate TacZero through experiments addressing four questions: (1) Can a pretrained general-purpose VLM infer contact conditions from visual and tactile observations and select actions to perform peg insertion without additional tactile or manipulation training or hand-designed task-specific rules? (2) Does tactile input improve insertion performance over a no-tactile baseline? (3) How does insertion performance differ between numerical and overlay representations of the tactile responses? (4) Can TacZero also perform insertion with another VLM and a different peg shape?

To facilitate reproduction of the setup, we use cylindrical and rectangular-prism pegs with their corresponding hole fixtures from the Functional Manipulation Benchmark (FMB)~\cite{luo2025fmb}. We select the Large size and Short length for both peg shapes. The hole fixture remains in the same location throughout each comparison, with a non-slip mat underneath to prevent movement. For cylindrical-peg insertion, the instruction is: ``Fully insert the purple cylinder into the central hole of the green fixture until it reaches the bottom. Maintain your grasp until insertion is complete.'' For rectangular-prism peg insertion, the instruction is: ``Fully insert the light blue rectangular prism into its corresponding hole until it reaches the bottom. Maintain your grasp until insertion is complete.''

Using GPT-6 Astra (\texttt{gpt-6-astra}), we compare three input conditions: a no-tactile baseline, which receives camera images and robot state without tactile input; TacZero-Numbers, which additionally receives aggregate three-axis tactile responses as numerical text; and TacZero-Overlay, which receives these responses as arrows overlaid on the camera images. We conduct 20 independent cylindrical-peg insertion trials per condition. Both tactile conditions use the same baseline-corrected three-axis responses, presented as numerical text or image overlays. We additionally evaluate TacZero-Numbers in 10 independent rectangular-prism peg insertion trials using Astra and in 10 independent cylindrical-peg insertion trials using Claude Fable 5.1 (\texttt{claude-fable-5-1}) at high reasoning effort. All Astra trials use medium reasoning effort. Across these experiments, we use the same robot-state inputs, translation controller, execution limits, and history-inclusion scheme, which retains all previous observation images from the same trial. The task instruction is the same for all trials with the same peg shape. The base prompt includes the instruction: ``Use exactly one tool each turn and include a short observation and reason in note.'' In the tactile conditions, we append ``considering tactile input when available'' to this instruction.

Before each trial, the experimenter returns the arm to its initial pose and manually places the peg in approximately the same starting region. To focus on TacZero's behavior during peg insertion, the robot follows a fixed sequence to grasp and lift the peg before VLM control begins. After the experimenter confirms the grasp, the VLM starts controlling insertion.

To examine whether a general-purpose VLM can infer contact conditions and select actions for peg insertion, we analyze the observations and reasons recorded in its notes alongside executed actions and visual and tactile observations across trials. To assess whether tactile input improves insertion performance and how its representation affects the results, we compare human-judged success rates and failure categories across the three input conditions. The experimenter assigns each failed trial to one of three mutually exclusive categories: no entry, jamming after entry, or grasp slippage. Here, grasp slippage refers to displacement of the peg relative to the gripper fingers caused by contact forces from the hole fixture. Finally, we report action counts, execution time, API waiting time, token usage, and estimated API cost.

\subsection{Results}

\begin{table}[t]
  \centering
  \caption{Action counts, execution time, token usage, and estimated API cost.}
  \label{tab:resources}
  \footnotesize
  \setlength{\tabcolsep}{2pt}
  \renewcommand{\arraystretch}{1.12}
  \begin{tabular*}{\linewidth}{@{\extracolsep{\fill}}lccccc@{}}
    \hline
    & None & Num. & Overlay & Box & Fable \\
    \hline
    Trials & 20 & 20 & 20 & 10 & 10 \\
    Actions & 19.9 & 24.3 & 23.2 & 23.0 & 8.9 \\
    \hline
    \multicolumn{6}{l}{\textit{Mean time per trial}} \\
    Motion (s) & 94 & 110 & 105 & 99 & 52 \\
    API (s) & 177 & 239 & 224 & 223 & 277 \\
    Time (s) & 271 & 349 & 329 & 322 & 330 \\
    \hline
    Tokens (M) & 0.952 & 1.817 & 1.731 & 1.544 & 0.419 \\
    \hline
    \multicolumn{6}{l}{\textit{Estimated cost (USD)}} \\
    Cost (\$) & 12.02 & 22.84 & 21.41 & 19.43 & 4.91 \\
    Total (\$) & 240.49 & 456.83 & 428.13 & 194.35 & 49.05 \\
    \hline
  \end{tabular*}
  \par\smallskip
  \parbox{\linewidth}{\footnotesize None, Num., and Overlay denote the no-tactile baseline, TacZero-Numbers, and TacZero-Overlay, respectively, for cylindrical-peg insertion with Astra. Box and Fable denote rectangular-prism insertion with Astra and cylindrical-peg insertion with Fable, respectively, both using TacZero-Numbers.}
\end{table}

Table~\ref{tab:resources} summarizes action counts, execution time, API waiting time, token usage, and estimated API cost. All values except the trial counts and total costs are means per trial. Time is the sum of action execution and API waiting, excluding preparation, image uploads, and other processing overhead. Token usage includes input and output tokens, and costs are estimated from published rates, excluding calls with unreported usage. API waiting time exceeded action execution time in all conditions. For cylindrical-peg insertion with Astra under the TacZero-Numbers condition, the combined action-execution and API-waiting time averaged 349~s per trial, of which 239~s was spent waiting for the API. The entire insertion phase, including image processing, uploads, and other overhead, averaged 446~s. Fable used fewer actions and had a lower estimated API cost. The estimated API cost across all 80 trials was USD 1,369.85.

\begin{figure}[t]
  \centering
  \includegraphics[width=\linewidth]{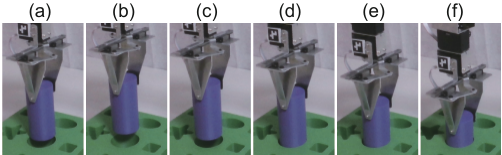}
  \caption{Successful TacZero-Numbers trial 3: (a) entrance contact, (b) retraction, (c) realignment, (d) entry, (e) insertion, and (f) completion report.}
  \label{fig:success}
\end{figure}

Fig.~\ref{fig:success} shows a successful cylindrical-peg insertion trial with TacZero-Numbers. At the beginning of the trial, the VLM described the peg as held upright beside the fixture and selected a higher target end-effector position, citing the need for clearance before lateral motion. Although the hole's numerical coordinates were not supplied, the VLM then selected a target end-effector position above the estimated hole center, citing the camera images and calibration information and noting that the lift had been visually confirmed. After approaching the hole, the VLM specified a higher target end-effector position. In its note, it described stalled downward motion and an increased tactile response in the vertical direction of the robot reference frame and gave entrance contact rather than bottom contact as the reason for retracting (a). It then specified a target end-effector position offset in XY above the opening. Its note stated that the images and TCP measurements indicated clearance and that the tactile response had subsided (b). Panel (c) shows a later entry attempt. After the peg entered the hole, the VLM specified lower target end-effector positions, citing visible insertion progress and the absence of an increase in the vertical tactile response in its notes (d, e). It then specified lower target positions in smaller increments and reported completion. In its final note, it described little visible change in insertion depth and an increased vertical tactile response, identifying bottom contact as the reason for reporting completion (f).

\begin{table}[t]
  \centering
  \caption{Insertion outcomes by condition.}
  \label{tab:insertion-results}
  \footnotesize
  \setlength{\tabcolsep}{2pt}
  \renewcommand{\arraystretch}{1.12}
  \begin{tabular*}{\linewidth}{@{\extracolsep{\fill}}lccccc@{}}
    \hline
    & None & Num. & Overlay & Box & Fable \\
    \hline
    Trials & 20 & 20 & 20 & 10 & 10 \\
    Successes & 10 & 15 & 11 & 5 & 2 \\
    Success (\%) & 50 & 75 & 55 & 50 & 20 \\
    \hline
    \multicolumn{6}{l}{\textit{Failures (trials)}} \\
    No entry & 2 & 2 & 3 & 2 & 7 \\
    Jamming & 6 & 3 & 6 & 3 & 0 \\
    Slip & 2 & 0 & 0 & 0 & 1 \\
    \hline
  \end{tabular*}
  \par\smallskip
  \parbox{\linewidth}{\footnotesize None, Num., and Overlay denote the no-tactile baseline, TacZero-Numbers, and TacZero-Overlay, respectively, for cylindrical-peg insertion with Astra. Box and Fable denote rectangular-prism insertion with Astra and cylindrical-peg insertion with Fable, respectively, both using TacZero-Numbers.}
\end{table}

Table~\ref{tab:insertion-results} presents the success rates. In cylindrical-peg insertion with Astra, TacZero-Numbers succeeded in 15 of 20 trials (75\%), compared with 10 of 20 (50\%) for the no-tactile baseline and 11 of 20 (55\%) with TacZero-Overlay. TacZero-Numbers yielded the highest observed success rate, whereas the success rate with TacZero-Overlay was close to that of the no-tactile baseline. These results suggest that tactile input can improve insertion performance, but its benefit may depend on how it is represented to the VLM.

TacZero-Numbers also performed successful insertions with another peg shape and another VLM. With Astra, it succeeded in 5 of 10 rectangular-prism peg insertion trials (50\%). With Fable, it succeeded in 2 of 10 cylindrical-peg insertion trials (20\%). The rectangular-prism failures comprised two trials without hole entry and three with jamming after entry. The Fable failures comprised seven trials without hole entry and one with grasp slippage.

In the cylindrical-peg comparison, failures to enter the hole occurred in 2 trials with the no-tactile baseline, 2 with TacZero-Numbers, and 3 with TacZero-Overlay. Failures due to jamming after entry occurred in 6, 3, and 6 trials, respectively. Two further failures with the no-tactile baseline were attributed to grasp slippage. Thus, TacZero-Numbers had fewer jamming failures, while failures to enter the hole were similar across conditions. In its notes, the VLM frequently referred to vertical tactile responses, whereas references to horizontal responses were limited even in trials judged by the experimenter to have failed due to jamming during insertion.

In both the TacZero-Numbers and TacZero-Overlay conditions, some failed trials ended before the peg entered the hole, with the VLM explicitly stating in its final note that the task was incomplete. In other failed trials in both conditions, the VLM stated in its final note that the peg had reached the bottom and reported completion, although the experimenter judged that the peg had jammed before reaching the required insertion depth.

\section{DISCUSSION}

In our experiments, TacZero-Numbers yielded a higher peg-insertion success rate than the no-tactile baseline. The general-purpose VLM retracted and reinserted the peg, citing tactile responses in its notes. These results show that TacZero can perform peg insertion by using a general-purpose VLM to interpret visual and tactile observations and select robot actions, without additional tactile or manipulation training or hand-designed rules for contact interpretation or action selection. TacZero provides a concrete baseline for developing and comparing future methods. The following limitations highlight challenges for further research on this approach.

One limitation concerns how we presented observations to the VLM. We used a single camera viewpoint, leaving some aspects of peg--hole alignment and contact visually ambiguous. In its notes, the VLM frequently referred to vertical tactile responses but infrequently referred to horizontal responses. This tendency may depend in part on the fixed end-effector orientation and the textual descriptions of coordinate-frame relationships provided in our setup. We compared numerical and overlay representations only within this setup, so the advantage of TacZero-Numbers over TacZero-Overlay may not extend to other viewpoints or coordinate descriptions. One direction for future work is therefore to jointly design camera viewpoints, coordinate descriptions, and tactile representations to help general-purpose VLMs better interpret visual and tactile observations and select robot actions.

A similar limitation concerns how we specified the goal and action constraints. We provided the goal in text without an image of the completed task, so the VLM could not compare its current view with a visual reference of completion. In some failed trials, the VLM reported completion even though the experimenter judged that the peg had not entered the hole or had jammed before reaching the required insertion depth. We left the decision to continue or terminate an attempt to the VLM, which sometimes abandoned attempts while further corrections remained possible. We also required the VLM to maintain its grasp, preventing regrasping after slippage. Another direction for future work is to design human--VLM interaction, including how goals and instructions are conveyed, so that the VLM's actions better reflect human intent.

Another limitation is the latency of VLM responses. In our implementation, each movement toward a target end-effector position at a fixed orientation was followed by an observation and a wait for the VLM's next command. As a result, increases in tactile responses or slippage of the peg relative to the gripper fingers during motion or while waiting for a response could not trigger immediate stopping or motion adjustments. Beyond movements toward individual target positions at a fixed end-effector orientation, future work should support a broader range of motion trajectories and changes in end-effector orientation. A further direction is to design a hierarchical system that asynchronously combines VLM-based contact interpretation and action selection with rapid monitoring and stopping and tactile feedback control.

\section{CONCLUSION}

We proposed TacZero, which uses a pretrained general-purpose VLM to interpret visual and tactile observations and select robot actions without additional tactile or manipulation training or task-specific rules for contact interpretation or action selection. In real-world cylindrical-peg insertion experiments, TacZero succeeded in 15 of 20 trials with numerical tactile input, compared with 10 of 20 without tactile input. This study provides a concrete starting point for further research on this approach and highlights challenges in pursuing this direction.

\section*{Acknowledgments}

The authors used OpenAI Codex with GPT-5.6 Sol and GPT-6 Astra to assist in drafting text throughout the manuscript and in developing experimental software that incorporated pre-existing code.

\bibliographystyle{IEEEtran}
\bibliography{reference}

\end{document}